\documentclass[letterpaper, 10 pt, conference]{ieeeconf}  

\IEEEoverridecommandlockouts                              
\usepackage{amsmath} 
\usepackage{amssymb}  
\usepackage{cite}
\usepackage{graphicx}


\title{\LARGE \bf
Deep Canonically Correlated LSTMs
}

\author{
	Mallinar, Neil\\
    \texttt{nmallinar@gmail.com}
    \and
    Rosset, Corbin\\
    \texttt{corbyrosset@gmail.com}
}

\begin{document}

\maketitle
\thispagestyle{empty}
\pagestyle{empty}

\begin{abstract}
We examine Deep Canonically Correlated LSTMs as a way to learn nonlinear transformations of variable length sequences and embed them into a correlated, fixed dimensional space. We use LSTMs to transform multi-view time-series data non-linearly while learning temporal relationships within the data. We then perform correlation analysis on the outputs of these neural networks to find a correlated subspace through which we get our final representation via projection. This work follows from previous work done on Deep Canonical Correlation (DCCA), in which deep feed-forward neural networks were used to learn nonlinear transformations of data while maximizing correlation.
\end{abstract}

\section{INTRODUCTION}

It is common in modern data sets to have multiple views of data collected of a phenomenon, for instance, a set of images and their captions in text, or audio and video data of the same event. If there exist labels, the views are conditionally uncorrelated on them, and it is typically assumed that noise sources between views are uncorrelated so that the representations are discriminating of the underlying semantic content. To distinguish it from multi-modal learning, multi-view learning trains a model or classifier for each view, the application of which depends on what data is available at test time. Typically it is desirable to find representations for each view that are predictive of - and predicted by - the other views so that if one view is not available at test time, it can serve to denoise the other views, or serve as a soft supervisor providing pseudo-labels. The benefits of training on multiple views include reduced sample complexity for prediction scenarios ~\cite{kakade2007multi}, relaxed separation conditions for clustering ~\cite{chaudhuri2009multi}, among others. CCA techniques are used successfully across a wide array of downstream tasks (often unsupervised) from fMRI analysis ~\cite{hardoon2007unsupervised}, to retrieval, categorization, and clustering of text documents ~\cite{vinokourov2002inferring,benton2016learning}, to acoustic feature learning ~\cite{bharadwaj2012multiview,arora2013multi,wang2015unsupervised}.

Canonical Correlation Analysis (CCA) is a widely used procedure motivated by the belief that multiple sources of information might ease learnability of useful, low-dimensional representations of data ~\cite{hotelling1935canonical}. Specifically, CCA learns linear projections of vectors from two views that are maximally correlated ~\cite{hotelling1936relations}. While CCA is affine invariant, the representations it learns are not sufficient for data that lies on a manifold. Kernel Canonical Correlation Analysis (KCCA) finds similar non-linear low dimensional projections of data \cite{akaho2006kernel}. We apply a scalable\footnote{Although the dimension of the XRMB data is on the order of $10^2$, the number of samples ($\approx$ 50,000) would force the Gram matrix to exceed memory capacity} extension of kernel CCA (KCCA) with Gaussian and polynomial kernels. However, kernel methods scale poorly in both feature dimension (a dot product or norm is required) and number of samples (Gram matrix is polynomial in size). 

To remedy this, we apply deep methods, which enjoy learning a parametric form of a nonlinear target transformation. One such method is that of split autoencoders (SplitAE)\footnote{As described later, SplitAE reconstructs each view from a shared representation, which is not correlation analysis}. This method was originally proposed by Ngiam \textit{et al.} in their paper on multimodal autoencoders \cite{ngiam2011multimodal} and was shown to be comparable to correlation-based methods by Wang \textit{et al.} in their survey of multi-view methods \cite{wang2015deep}. Andrew \textit{et al.} proposed Deep Canonical Correlation Analysis (DCCA) as a deep correlation-based model for multi-view representation learning \cite{andrew2013deep}. DCCA maps various views of data into a lower dimension via deep feed-forward neural networks and performs linear CCA on the output of these networks to learn deep correlation representations of data in an unsupervised fashion. The obvious extension to this method was to include the reconstruction objective of multimodal autoencoders with the correlation objective to get Deep Canonically Correlated Auto-Encoders (DCCAE), which outperformed standard DCCA \cite{wang2015deep}.

However, we note that these deep methods for correlation analysis require using fixed length feature vectors, which removing the long-term time dimension from our data. In this paper, we explore the use of Long Short Term Memory (LSTM) ~\cite{hochreiter1997long} cells to learn temporal relationships within our sequences. Recurrent Neural Networks (RNN's) have been shown to work very well with time-series data, including but not limited to audio and video sequences. LSTM cells are extensions of standard RNNs that allow for the network to persist what it learns over longer periods of time ~\cite{gers2001long}. These networks have been shown to perform exceptionally well at tasks such as speech recognition ~\cite{hannun2014deep}, ~\cite{sak2014long}, acoustic modeling ~\cite{sak2014long2}, sentence embedding ~\cite{palangi2015deep} and more.

We present results of a baseline LSTM classifier and how well it clusters phonemes, as well as baseline linear CCA with a classifier on the projected data, and finally DCC-LSTM with a classifier on the final projected data. We then compare these results to other deep methods such as SplitAE and DCCAE.


\section{Background and Related Work}

Here we will review past work done in correlation analysis, deep methods for multi-view representation learning, and LSTMs in order to provide a solid theoretical justification for the methods we are using.

\subsection{Canonical Correlation Analysis}

CCA can be interpreted as an extension of principle component analysis to multiple data sets with the added constraint that the principle components learned in each subspace are maximally correlated. Concretely, given $n$ observations $(x_i, y_i)$,  $x_i \in \mathbb{R}^{d_x}$ and $y_i \in \mathbb{R}^{d_y}$ comprising two views of data $\mathcal{X}$, $\mathcal{Y}$ described by an unknown joint distribution $\mathcal{D}$, find $k$ pairs of vectors $(u_j, v_j)$ of the same dimensions to 
\begin{equation}
\text{max } correlation(u_i^{\top}x, v_i^{\top}y)
\label{eqn:lincorrobj}
\end{equation}
subject to the constraint that $(u_j, v_j)$ is uncorrelated to all other $(u_r, v_r)$, $j \neq r$, $1 \leq j \leq k$. After finding the first pair $(u_1, v_1)$, subsequent pairs are found via deflation of the views subject to the uncorrelation constraint above. Expanding the definition of correlation for the vectors $u_i$ and $v_i$:
\begin{equation}
\begin{aligned}
& \underset{u_i \in \mathbb{R}^{d_x}, v_i \in \mathbb{R}^{d_y}}{\text{max}}
& & \frac{\mathbb{E}_{(x, y) \sim \mathcal{D}}\left[u_i^{\top}xy^{\top}v_i\right]}{\sqrt{\mathbb{E}_x[u_i^{\top}xx^{\top}u_i] \mathbb{E}_y[v_i^{\top}yy^{\top}v_i] }} \\
\end{aligned}
\label{eqn:lincorrexp}
\end{equation}
Note that scaling of $u_i$ or $v_i$ does not change the objective, therefore the variance of $u$ and $v$ can be normalized to one. Expressed equivalently for all $u$ and $v$ to be learned, 
\begin{equation}
\begin{aligned}
\underset{U \in \mathbb{R}^{d_x \times k}, V \in \mathbb{R}^{d_y \times k}}{\text{max}} & \mathbb{E}_{(x, y) \sim \mathcal{D}}\left[\text{trace}\left(U^{\top}xy^{\top}V\right)\right] \\
\text{subject to \ \ \ } & \mathbb{E}[U^{\top}xx^{\top}U] = I \\ 
& \mathbb{E}[V^{\top}yy^{\top}V] = I
\end{aligned}
\label{eqn:lincorrtr}
\end{equation}
Solving for $u$ and $v$ using the Lagrange multiplier method, linear CCA takes the form of a generalized eigenvalue problem, for which the solution is a product of covariance\footnote{For numerical stability, a scaled identity matrix is added to a covariance matrix before it is inverted} and cross-covariance matrices ~\cite{rastogi2015multiview}: $U$ is the top $k$ left eigenvectors (sorted by decreasing eigenvalue) of $C_{xx}^{-1}C_{xy}C_{yy}^{-1}C_{yx}$, which when regularized becomes $(C_{xx} + r_xI)^{-1/2}C_{xy}(C_{yy} + r_yI)^{-1/2}C_{yx}$. The $i$'th column of $V$ is then chosen as $\cfrac{C_{yy}^{-1}C_{yx}u_i}{\sqrt{\lambda_i}}$, which is regularized similarly. 

A second interpretation of CCA is that optimal $U$ and $V$ projection matrices minimize the squared error of reconstructing $x$ (respectively, $y$) given $y$ (resp. $x$). Hence, the optimization problems given in Equations \ref{eqn:lincorrobj}, \ref{eqn:lincorrexp}, \ref{eqn:lincorrtr}, and \ref{eqn:lincorrreconstruct} are all equivalent (for $i$ = 1...$k$, where applicable).
\begin{equation}
\begin{aligned}
\underset{U \in \mathbb{R}^{d_x \times k}, V \in \mathbb{R}^{d_y \times k}}{\text{min}} & \mathbb{E}_{(x, y) \sim \mathcal{D}}\left[ \|U^{\top}x - V^{\top}y\|_2^2 \right] \\
\text{subject to \ \ \ } & \mathbb{E}[U^{\top}C_{xx}U] = I_k \\
&\mathbb{E}[V^{\top}C_{yy}V] = I_k
\label{eqn:lincorrreconstruct}
\end{aligned}
\end{equation}
There are a number of settings to which CCA, and all correlation analysis variants, can be applied. The first is when all views are available at test and train time. The second is when only a non-empty subset of the views is available at test time. Either of these settings can be enhanced by labels. 

\subsection{Kernel Canonical Correlation Analysis}

KCCA by Akaho \textit{et al.} generalizes the transformations learned by CCA to nonlinear functions living in a reproducing kernel hilbert space (RKHS), making it suitable for manifold learning. Given $\mathcal{F}$ and $\mathcal{G}$ are function classes living in reproducing kernel hilbert spaces reserved for $\mathcal{X}$ and $\mathcal{Y}$ respectively, the goal is to find two sets of $k$ functions, $\{f_1, ..., f_k\} : \mathbb{R}^{d_x} \rightarrow \mathbb{R}$, $\{g_1, ..., g_k\} : \mathbb{R}^{d_y} \rightarrow \mathbb{R}$ such that for any $i$, $f_i$ and $g_i$ minimize the squared error of reconstructing each other in the RKHS.  That is, given $x \in \mathcal{X}$ and $y \in \mathcal{Y}$, $f$ and $g$ are maximally predictive of, and predictable by, the other, 
\begin{equation}
\begin{aligned}
& \underset{U \in \mathbb{R}^{d_x \times k}, V \in \mathbb{R}^{d_y \times k}}{\text{min}}
& \mathbb{E}_{(x, y) \sim \mathcal{D}}\left[ \|f_i(x) - g_i(y)\|_2^2 \right] \\
& \text{subject to}
& \mathbb{E}[f_i(x)f_j(x)] = \delta_{ij}, \\
& & \mathbb{E}[g_i(y)g_j(y)] = \delta{ij}
\end{aligned}
\label{eqn:kccarec}
\end{equation}
for $\delta_{ij}$ an indicator random variable that assumes one if $i = j$ and zero otherwise. We will also introduce equivalent interpretations similar to the above: maximize the correlation of $f(x)$ and $g(x)$ or maximize the covariance $\mathbb{E}_{(x, y) \sim \mathcal{D}}\left[ f_i(x)g_i(y)\right]$ with the same constraints as Equation \ref{eqn:kccarec}. As before, every pair of functions $(f_j, g_j)$ is uncorrelated with all other pairs of functions for all $k$.

For $n$ data vectors from each of $\mathcal{X}$ and $\mathcal{Y}$, the Gram matrices $K_x$ and $K_y$ from $\mathbb{R}^{n \times n}$ can be constructed\footnote{e.g. $[K_x]_{ij} = \kappa(x_i, x_j)$. Note also that the matrices are assumed to be centered} given a positive definite kernel function $\kappa( \cdot, \cdot) : \mathcal{X} \times \mathcal{Y} \rightarrow \mathbb{R}$ which is by construction equal to the dot product of its arguments in the feature space\footnote{known as the kernel trick}, $k(u, v) = \Phi(u)^{\top}\Phi(v)$ for a nonlinear feature map $\Phi : \mathbb{R}^d \rightarrow \mathcal{H}$ . It is a property of an RKHS that each of its constituent functions $f$ and $g$ from each view can be represented as a linear combination of the $n$ observed feature vectors, $f_i(x) = \sum_{i = 1}^n \alpha_j^{(i)}\kappa(x, x_j) = \alpha K_x(x)$. Similarly, $g_i(y) = \beta K_y(y)$. Mirroring arguments from CCA\footnote{An unconstrained KCCA objective can be written by moving the square root of the product of the constraints in Equation 6 to the denominator and removing the trace operator as in Equation 2.}, minimizing the objective in Equation 5 is equivalent to maximizing $\text{trace} \left( \alpha^{\top}K_xK_y\beta\right)$ where $\mathbb{E}\left[ f_i(x)f_j(x) \right] = \frac{1}{n}\alpha_i^{\top}K_x^2\alpha_j$. Absorbing the $\frac{1}{n}$ into $\alpha$, Equation \ref{eqn:kccarec} can be interpreted as
\begin{equation}
\begin{aligned}
& \underset{\alpha \in \mathbb{R}^{n \times k}, \beta \in \mathbb{R}^{n \times k}}{\text{max}}
& \text{trace } \alpha^{\top}K_xK_y\beta \\
& \text{subject to}
& \alpha^{\top}K_x^2\alpha = I_k, \\
& & \beta^{\top}K_y^2\beta = I_k
\end{aligned}
\label{eqn:kccatr}
\end{equation}
As before, $\alpha$ can be solved as a generalized eigenvalue problem, whose regularized solution is $\alpha = K_x^{-1}K_yK_y^{-1}K_x$ and the $i$th $\beta$ vector is $\beta_{(i)} = \frac{1}{\lambda_i}(K_y + r_yI)^{-1}K_x\alpha$. This solution takes $O(n^3)$ time and $O(n^2)$ space. When $k << n$, it is wasteful to compute all $n$ columns of $\alpha$ and $\beta$. Scalable implementations of KCCA employ rank-$m$ approximations of the eigendecompositions that give the solutions to $\alpha$ and $\beta$.

The primary drawback of KCCA is its nonparametric form, which requires $\alpha$ to be stored for use on all unseen examples (test sets).  Also, computation of dot products for the kernel functions cannot be avoided; $O(n^2)$ of these dot products are needed at training time. The binding of each instance of KCCA to a specific kernel also limits the function class of our solution. In high dimensional settings, KCCA is also susceptible to overfitting, that is, learning spurious correlations. A remedy is to regularize with a scaled identity matrix those Gram matrices to be inverted as in Equation \ref{eqn:kccareg}. The regularization parameters also need to be tuned. 
\begin{equation}
\begin{aligned}
& \underset{\alpha \in \mathbb{R}^{n \times k}, \beta \in \mathbb{R}^{n \times k}}{\text{max}}
& & \frac{ \alpha^{\top}K_xK_y\beta}{\sqrt{K_x^{(reg)}K_y^{(reg)}}} \\
\end{aligned}
\label{eqn:kccareg}
\end{equation}
where
\begin{equation}
\nonumber
\begin{aligned}
K_x^{(reg)} = \alpha^{\top}K_x^2\alpha + r_x\alpha^{\top}K_x\alpha \\
K_y^{(reg)} = \beta^{\top}K_y^2\beta + r_y\beta^{\top}K_y\beta
\end{aligned}
\end{equation}
The regularized solution for $\alpha$ is the top eigenvectors sorted by decreasing eigenvalue of the matrix $(k_x + r_xI)^{-1}K_y(K_y + r_yI)^{-1}K_x$ and the $i$th $\beta$ vector is as before. A scalable approach to kernel CCA was presented in~\cite{arora2012kernel}.

\subsection{Split AutoEncoders}

The work of Ngiam \textit{et al.} on multimodal autoencoders introduced a deep network architecture for learning a single representation from one or more views of data \cite{ngiam2011multimodal}. This was later followed up by Wang \textit{et al.} with applications to the Wisconsin X-Ray dataset, as well as a constructed multi-view version of the MNIST data set \cite{wang2015deep}. The goal of deep autoencoder architectures in multiview learning is to find some shared representation that minimizes the reconstruction error for all views. These deep autoencoders can take as input one or many views at the same time, depending on which are available for testing. For the sake of this paper, we will restrict ourselves to two views.

There are two architectures given by Ngiam \textit{et al.} that find a shared representation of two views ~\cite{ngiam2011multimodal}. The dataset used to train these have two views available at train time and one view available at test time, as is the case with the XRMB dataset that we are using in this paper.
\begin{equation}
\min_{\mathbf{W}_f, \mathbf{W}_g, \mathbf{W}_p, \mathbf{W}_q} \frac{1}{2} \sum\limits_{i=1}^N (\Vert \mathbf{x}_i - \mathbf{p}(\mathbf{f}(\mathbf{x}_i)) \Vert^2 + \Vert \mathbf{y}_i - \mathbf{q}(\mathbf{g}(\mathbf{y}_i)) \Vert^2)
\label{eqn:splitAEtwotrain}
\end{equation}
The first architecture trains using information from both views, minimizing Equation \ref{eqn:splitAEtwotrain} which is the sum of the $\text{L}_2$ Norm of the reconstruction for both views. Both inputs are fed into the autoencoder separately and go through multiple hidden layers before being combined into a shared hidden layer representation of the two views. The decoders are then symmetrical to the encoders. At test time, all of the weights from the decoder of the view not available at test time are ignored, and the shared hidden layer representation calculated from the single view available is used.
\begin{equation}
\min_{\mathbf{W}_f, \mathbf{W}_p, \mathbf{W}_q} \frac{1}{2} \sum\limits_{i=1}^N (\Vert \mathbf{x}_i - \mathbf{p}(\mathbf{f}(\mathbf{x}_i)) \Vert^2 + \Vert \mathbf{y}_i - \mathbf{q}(\mathbf{f}(\mathbf{x}_i)) \Vert^2)
\label{eqn:splitAEonetrain}
\end{equation}
The second architecture, used by Wang et al, has a single encoder that takes as input the view available at train time. It then attempts to learn a shared representation and sets of weights that can reconstruct both views. The decoder for the view available at test time is symmetric to the encoder. The decoder for the view that's only available at train time can be a multilayer decoder or a single layer decoder that is experimentally tuned for number of layers and nodes.

\subsection{Deep Canonical Correlation Analysis}

Deep CCA, introduced by Andrew \textit{et al.}, is a parametric technique to simultaneously learn nonlinear mappings for each view which are maximally correlated. It is similar to KCCA in that both are computing nonlinear mappings to maximize canonical correlation between views. However, KCCA has a significant cost in that KCCA requires large kernel matrices which may not often be practical. DCCA on the other hand computes the nonlinear mappings using deep neural networks, which are capable of representing nonlinear, high-level abstractions on top of the input data. The use of neural networks makes DCCA much more scalable than standard KCCA, as the size of the network is not tied to the size of the dataset.

Given views $X$ and $Y$, DCCA learns representations $F$ and $G$ such that $F$=$f(X)$ and $G$=$g(Y)$ where $f$ and $g$ are the transformations computed by two deep neural networks, which are described by network weights $\mathbf{w_f}$ and $\mathbf{w_g}$ respectively. DCCA trains $f$ and $g$ according to the following objective, which maximizes the canonical correlation at the output layer between the two views.
\begin{equation}
\begin{aligned}
\max_{U, V, \mathbf{w_f}, \mathbf{w_g}} & \frac{1}{N} \mathbf{tr}(U^T F G^T V) \\
\text{subject to}\\
& U^T (\frac{F F^T}{N} + r_x I) U = I\\
& V^T (\frac{G G^T}{N} + r_y I) V = I
\end{aligned}
\end{equation}
An alternative objective, that we use in training, for DCCA can be expressed via the centered covariance matrices of the view 1 and view 2 data, $\hat{\Sigma}_{11}$ and $\hat{\Sigma}_{22}$, and the centered cross-covariance matrices from the two views, $\hat{\Sigma}_{12}$ and $\hat{\Sigma}_{21}$. 
We note first that DCCA must be trained in minibatches as there is no incremental CCA that is stable in this context and so we let $m$ be the batch size and center the transformed data batches via: $\bar{F} = F - \frac{1}{m}F\boldmath{1}_{m \times m}$ (resp. $G$). 
Define $\hat{\Sigma}_{11} = \frac{1}{m-1}\bar{F}\bar{F}^T + r_x I$ (resp. $\hat{\Sigma}_{22}$). Here we take $r_x, r_y > 0$ to be regularization parameters for the covariance matrices of $F, G$ respectively.
Then define $\hat{\Sigma}_{12} = \frac{1}{m-1}\bar{F}\bar{G}^T$.
From Andrew \textit{et al.} we define the matrix $T = \hat{\Sigma}_{11}^{-1/2}\hat{\Sigma}_{12}\hat{\Sigma}_{22}^{-1/2}$.
If the number of correlation components, $k$, that we seek to capture is the same as the output dimensionality, $o$, of the neural networks $f, g$ then we can express correlation as
\begin{equation}
\text{corr}(F, G) = \text{tr}(T^TT)^{1/2} = \text{tr}(\hat{\Sigma}_{21}\hat{\Sigma}_{11}^{-1}\hat{\Sigma}_{12}\hat{\Sigma}_{22}^{-1})^{1/2}
\label{eqn:correlation-objective}
\end{equation}
which is equivalent to the sum of the top $k$ singular values of the matrix $T$.

\begin{figure*}[t!]
\centering
\includegraphics[keepaspectratio, width=0.4\textwidth]{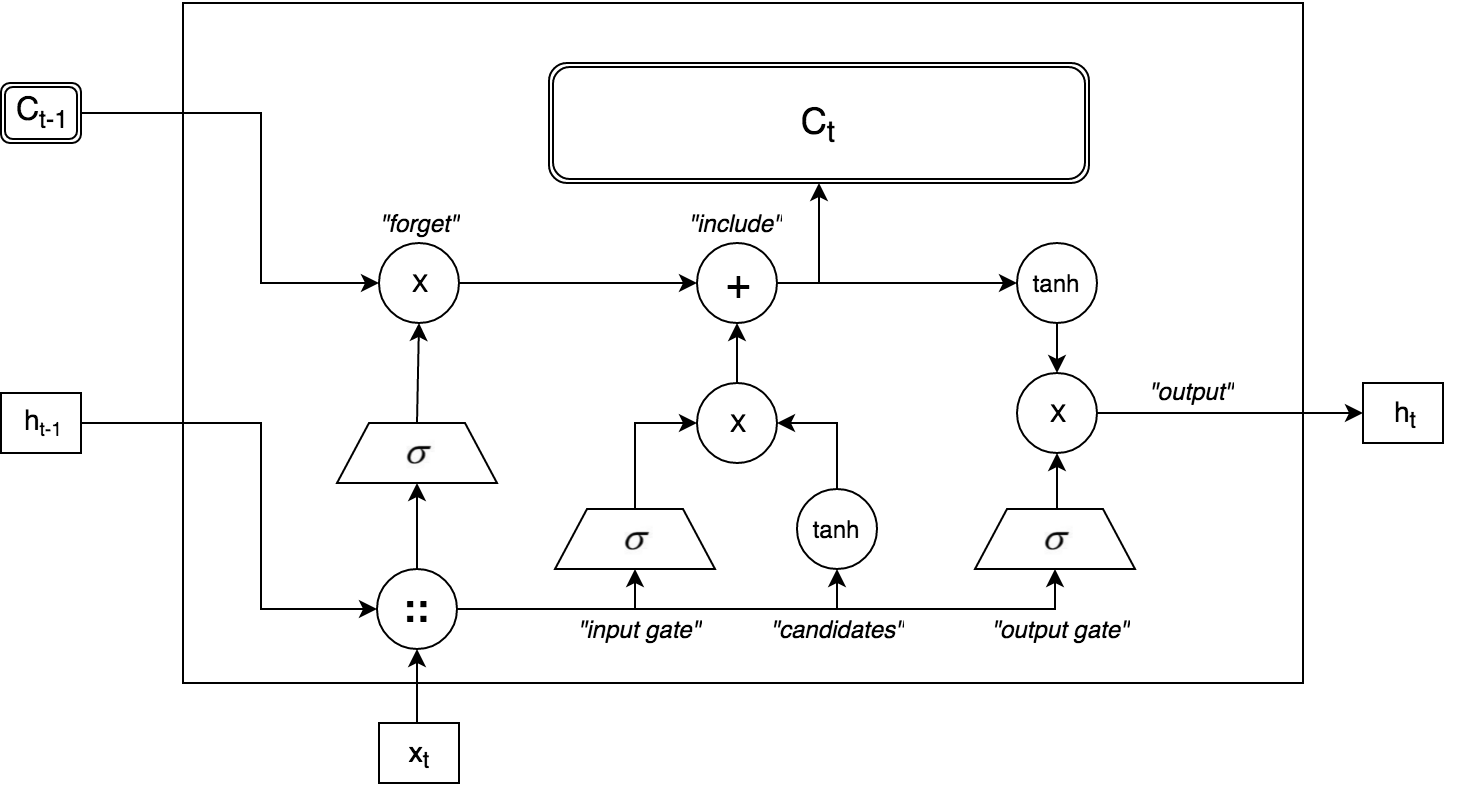}
\hspace{2em}
\includegraphics[keepaspectratio, width=0.4\textwidth]{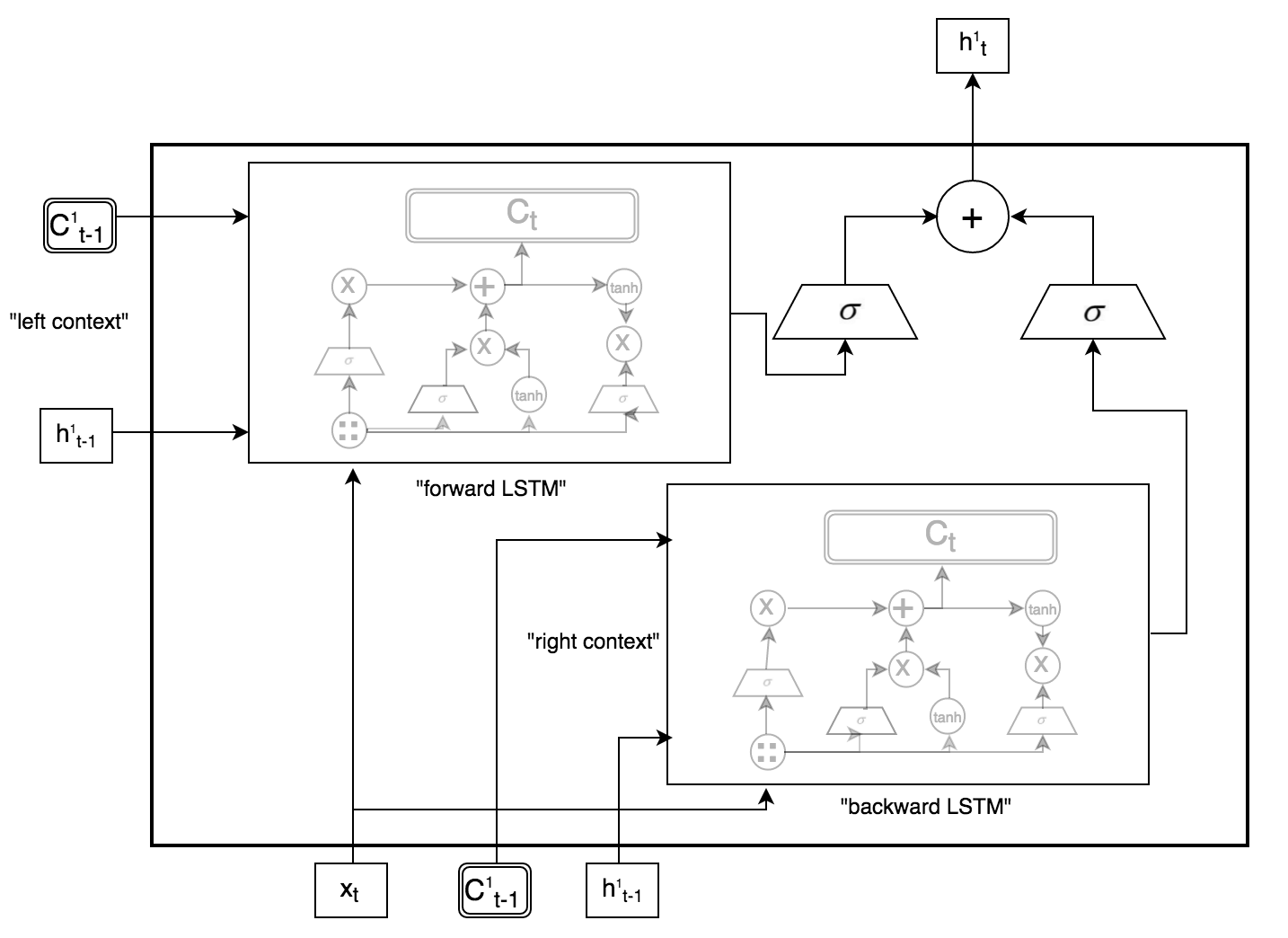}
\caption{Schematic of the traditional LSTM cell (forward and bidirectional cell variants), including all element-wise operations, inputs, outputs, and updates.}
\label{fig:lstm-cell}
\end{figure*}

\subsection{Long Short Term Memory Cell}

Long Short Term Memory (LSTM) Cells came out of recognition of many of the problems with recurrent neural networks (RNN). As such, it is a variant of the RNN architecture that solves many of these problems. The purpose of such recurrent neural networks is to handle time-series data in a more natural way, e.g. learning patterns across time. This is called sequence learning and can be seen as doing sequence-to-sequence or sequence-to-fixed dimensional learning.

The traditional LSTM cell consists of four networks that are trained in parallel. These four networks are: the forget gate, the input gate, the cell state, and the output. Of particular note is the cell state, which forms a "constant error carousel." The cell state avoids the problem of vanishing gradients and forward activation exploding because it is not updated with a gradient and is controlled by the forget and input gates (which are $[0, 1]$ bounded). ~\cite{gers2001long}.

A common variant on LSTMs is to add a peephole connction, in which you allow for the four aforementioned gates to look at the previous cell state (as opposed to just the previous time steps output and the current time steps input). Another common extension is to use bidirectional LSTMs instead of simply forward LSTMs. In the forward LSTM cell all of the past time steps influence the current one being evaluated; however, in the bidirectional LSTM cell all of the past and future time steps influence the current step being evaluated.

We will now give a more formal treatment of the LSTM cell, using Figure \ref{fig:lstm-cell} for reference. To update the peephole LSTM cell, we must evaluate
\begin{align*}
f_t &= \sigma(W_f[c_{t-1}, h_{t-1}, x_t] + b_f)\\
i_t &= \sigma(W_i[c_{t-1}, h_{t-1}, x_t] + b_i)\\
g_t &= \tanh(W_g[h_{t-1}, x_t] + b_g)\\
c_t &= f_t \times c_{t-1} + i_t \times g_t\\
o_t &= \sigma(W_o[c_{t-1}, h_{t-1}, x_t] + b_o)\\
h_t &= o_t \times \tanh(c_t)
\label{eqn:lstm-updates}
\end{align*}
at every time step.

The input to an LSTM cell is $[c_{t-1}, h_{t-1}, x_t]$ where $c_{t-1}$ is the cell state at the previous time step, $h_{t-1}$ is the output representation of the data at the previous time step, and $x_t$ is the input data at the current time step. This is then pass through a two separate sigmoid networks in order to get the forget and input gates. 

The forget gate, $f_t$, is $[0, 1]$ bounded. This represents what information from the past we want to forget now, where a value of 0 would mean to completely forget any given piece of information (element of the cell state) and a value of 1 would mean to not forget. The input gate, $i_t$, operates identically and is also $[0, 1]$ bounded, but it represents what new information we want to learn.

Next comes the new candidate values, $g_t$, which are the raw updates to the cell state. The final step, in order to update the cell state to the current time step, is to perform an element-wise multiplication of the previous cell state with the output of the forget gate and similarly the new candidate values with the output of the input game. Lastly, an element-wise addition is applied to get the new cell state.

The input to the cell is fed through a final sigmoid layer in order to get the output gate layer, $o_t$, which represents the raw cell output values. This is the element-wise multiplied with the cell state (under the tanh operator) in order to modulate the output with the information we've learned/forgotten.

The bidirectional LSTM cell operates using two forward LSTMs that traverse the data in opposite directions and is updated via
\begin{align*}
h_t^f &= \text{LSTM}_f(x_t, h_{t-1}^f)\\
h_t^b &= \text{LSTM}_b(x_t, h_{t-1}^b)\\
h_t &= W_fh_t^f + W_bh_t^b + b
\end{align*}
in which $\text{LSTM}_f$ and $\text{LSTM}_b$ are the forward and backward LSTMs, and the final output representation at the given time step is a weighted linear combination of the two cells outputs.

\begin{figure*}[t!]
\centering
\includegraphics[keepaspectratio, width=0.5\textwidth]{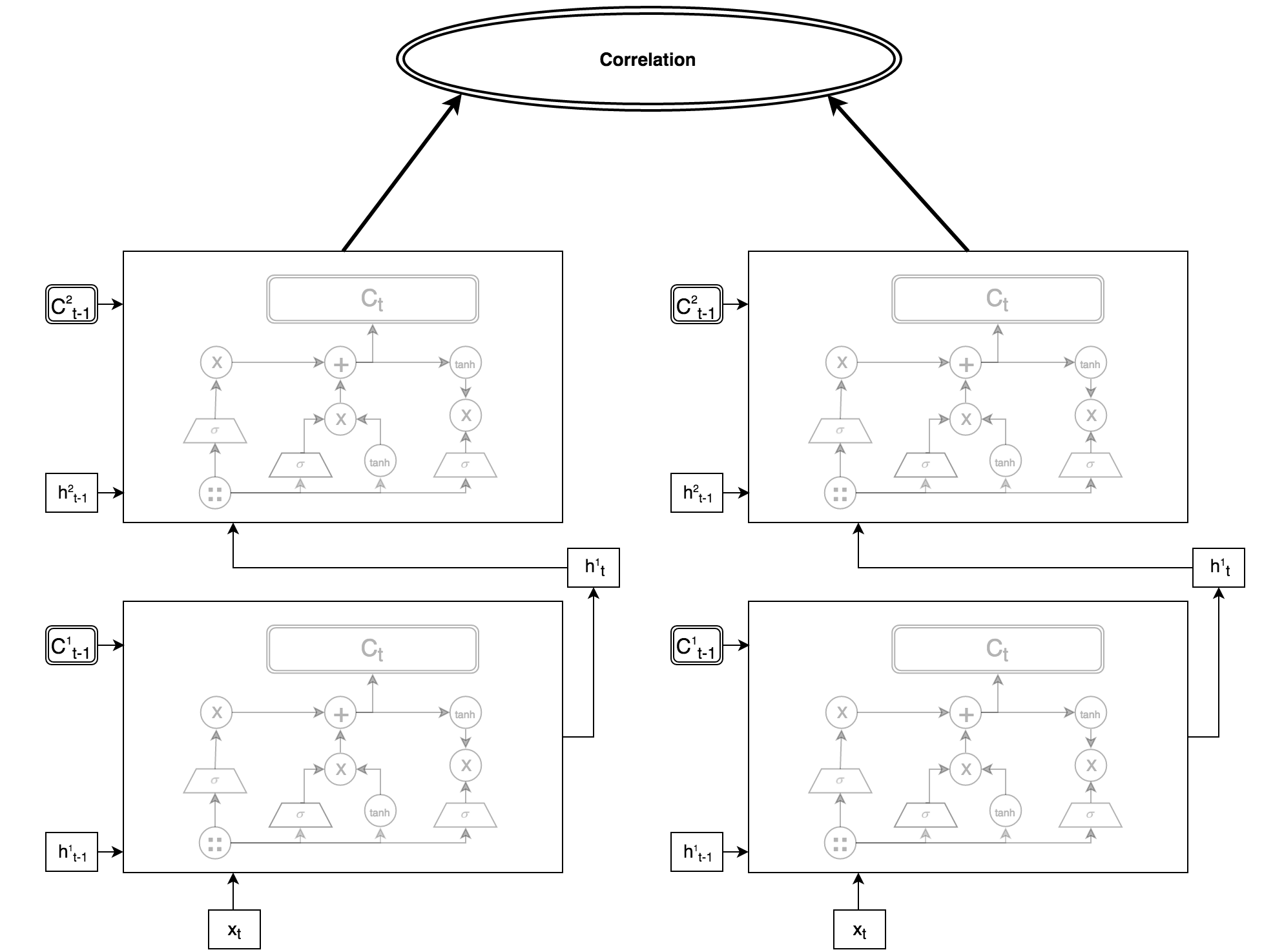}
\caption{DCC-LSTM schematic in which two time-series views are fed through deep LSTM networks and optimized using the correlation objective at the top.}
\end{figure*}

\section{Deep Canonically Correlated LSTMs}

In this paper we show the use of training deep LSTMs using correlation as an unsupervised objective. The Deep Canonically Correlated LSTM (DCC-LSTM) computes lower dimensional representations of (time-series) data using deep LSTM (DLSTM) networks on each view of the data. The final representations are fixed at the top layer of the DLSTMs and a linear CCA objective is maximized on the transformed data. We restrict ourselves to two views for the sake of this paper and the data we are working with, but the natural extension for more than two views would be to use Generalized Canonical Correlation Analysis (gCCA) on the output layers of the networks. ~\cite{tenenhaus2011regularized}.

\subsection{Implementation}

One goal of this paper was to implement the correlation object (and therefore DCCA as well as DCC-LSTM) using a more accessible framework, such as Tensorflow and Theano. There is one implementation of DCCA available in Theano, that we are aware of, but it contains numerous mistakes in the source code as well in the mathematics of the objective function (as expressed in Equation \ref{eqn:correlation-objective}).

Our first attempt at implementing the correlation objective was in Tensorflow. We ran into problems when trying to compute the matrix square root, as there is no Tensorflow operation for it. We attempted to compute the matrix square root as
\begin{align*}
A &= VDV^T \\
A^{1/2} &= VD^{1/2}V^T
\end{align*}
where $D^{1/2}$ is the element-wise square root of the eigenvalues of $A$. However, Tensorflow does not have a gradient defined for the eigenvalue decomposition operation and so without manually defining such a gradient or defining a new Tensorflow operation we were unable to implement the correlation objective. 

The alternative method would be to break outside of the computation graph generated by Tensorflow, compute linear CCA manually, and use those projections in the objective function. However, this would cause the backpropagation to be disconnected from the actual neural nets that generated the output representations. This is not desirable because we would have to manually specify the gradient values at each layer, and backpropogate them, rather than defer to Tensorflows symbolic differentiation routine (as we had hoped, in order to have a cleaner implementation of the correlation objective).

Our second attempt at implementing the correlation objective was in Theano. We were able to successfully define the matrix square root as specific above, and defer to Theanos computation graph and symbolic differentiation to handle the updates to the neural networks in an optimal fashion.

It is worth nothing that the function ``theano.scan'' is used to achieve the recurrence relation for each time step of the LSTM, and is not well optimized for the GPU as opposed to the CPU. In addition, multiple calls to scan per update results in significant overhead that causes DCC-LSTM to take roughly 2-3 times longer to train than DCCA.

\subsection{Training and Tuning}

Correlation is a population objective and thus training using minibatches provides a sample correlation value that is not the true population correlation. However, if two assumptions are held then we get a bound on the error between the true population correlation and the sample correlation. This bound shows us that we can train DCCA and DCC-LSTM using minibatches and stochastic gradient descent, rather than approximation methods such as (limited-memory) BFGS or full batch learning~\cite{2016arXiv160201024W}.  

In order to properly train DCCA using minibatches, we must first ensure that the data instances used to construct the minibatch are sampled independently and are identically distributed (i.i.d.). Secondly, we must ensure there is an upper bound on the magnitude of the neural networks. This is achieved by using nonlinear activation functions that are bounded, such as the sigmoid function or hyperbolic tangent function. Given these constraints the difference between the sample correlation and population correlation, in expectation, is bounded by a function of the regularization parameters chosen and the minibatch size. As expected, a larger minibatch size reduces the error.

For DCC-LSTM the same bounds hold as the LSTM cell uses only hyperbolic tangent and sigmoid activation functions, which are both bounded. In addition, we construct our batches by sampling i.i.d. sequences. This allows us to train DCC-LSTM using batch gradient descent (BGD), or other variants on BGD (such as ADAM, Adadelta, etc.).

LSTMs are not susceptible to vanishing gradients; however, they are sensitive to exploding gradients. During training, we have found several ways around this problem. The best methods to deal with this are: gradient clipping (find bounds on the value of the gradient experimentally); random normal initialization of the weight matrices $\in [0.1, 0.1]$; initialize weights with orthogonal matrices (eigenvalues lie on unit circle and determinant is $\pm 1$ helps with numerical stability).

LSTMs are trained using backpropagation through time (BPTT) or truncated BPTT (TBPTT). To perform BPTT, each layer of the LSTM (single LSTM cell) is unfolded through the number of time steps, $n$, it is given to create a linear chain of LSTM cells that represents the same computation as the loop in a single LSTM cell. After evaluation, gradients are calculated at each step along the unfolded network. Finally, all of these gradients are averaged and each cell is updated with this same average. TBPTT operates the same but rather than averaging gradients over all $n$ time steps, it only averages over the last $t < n$ time steps.

The process of BPTT (and TBPTT) is handled through Theano when calling the symbolic differentiation routines on the ``theano.scan'' function that we discussed above. This is often where the slow-down that we discussed occurs, in the unfolding and averaging (especially when sequence lengths are long, as desired for training LSTMs).

Now, when sampling i.i.d. sequences we have the option of using variable length sequences or fixed length sequences. The final representation that goes into the minibatch is the output from the last LSTM cell after feeding in the sequence (which contains information about the previous $n$ time steps). After randomly sampling $m$ such sequences, we use their outputs as the minibatch and calculate correlation at the top layer. We then perform BPTT at each layer of the LSTMs, with respect to both views.

The power of LSTMs can really be seen in this aspect of giving it a variable length sequence and receiving a fixed dimensional representation of that sequence. This is particularly powerful for the problem of speech recognition, which we study in this paper, as phoneme boundaries are usually not fixed and LSTMs can learn these boundaries via training with variable length sequences (in which the sequence length is also sampled randomly).

\section{Experiments}

Other than the correlation captured, we evaluated the quality of the representations of the two views of data learned by the DCC-LSTM using a few common downstream tasks: reconstruction of one view's feature from the other, sequence labeling, and classification. 

\subsection{Data}

We use the Wisconsin X-Ray Microbeam (XRMB) dataset for our study. This dataset consists of fifty two speakers, and each speaker contains two views. The first view consists of thirteen mel-frequency cepstral coefficient (MFCC) features and their first and second derivatives. These features are hand-crafted based on the raw audio signal, and the thirteen MFCCs are calculated in windows of 10ms each. The thirteen MFCCs and first and second derivatives then form a thirty-nine dimensional vector for a given 10ms frame of audio. Each instance in this view is a single frame ($\in \mathbb{R}^{39}$). 

The second view consists of articulatory measurements retrieved from eight pellets placed on the speakers lips, tongue, and jaw. Each pellet is measured for vertical and horizontal displacement as the person speaks. So for each 10ms frame, the second view consists of sixteen total measurements ($\in \mathbb{R}^{16}$).

As each phoneme is not contained wholly in 10ms of audio, we stack seven frames for context. Therefore there is a central frame which contains the label of the phoneme and three frames of left context and three frames of right context. So our MFCC feature vectors are $x_1 \in \mathbb{R}^{273}$ and our articulatory feature vectors are $x_2 \in \mathbb{R}^{112}$.

For the sake of this paper, we subsample the speakers for quicker train times. We use the first four speakers\footnote{JW\{11, 12, 13, 14\}} for unsupervised training. The test speakers\footnote{JW\{18, 29, 33, 34, 36, 42, 45, 48, 51, 53, 58, 60\}}, referred to as "downstream speakers", were held out. Their data was transformed into the feature space using the parameters learned during training. We then use the downstream speakers to evaluate additional metrics.

\begin{table}[]
\centering
\begin{tabular}{ll}
\hline \\
\textbf{Method} & \textbf{\% Test Accuracy} \\ \\ \hline \\
Baseline K-NN   & 79.70                     \\
CCA             & 83.20                         \\
KCCA            & 77.60                         \\
SplitAE         & 84.58                         \\ \hline
\end{tabular}
\caption{Accuracy achieved with best parameters for each correlation analysis method implemented. For CCA and KCCA, the best $k$ was found to 60, and the number of neighbors for K-NN was 4. The size of the output layer of splitAE was 50. The authors believe that KCCA requires more tuning.}
\label{tbl:classification}
\end{table}

\subsection{Classification and Clustering}

\begin{figure*}[t!]
\centering
\includegraphics[height=2in, width=2.3in]{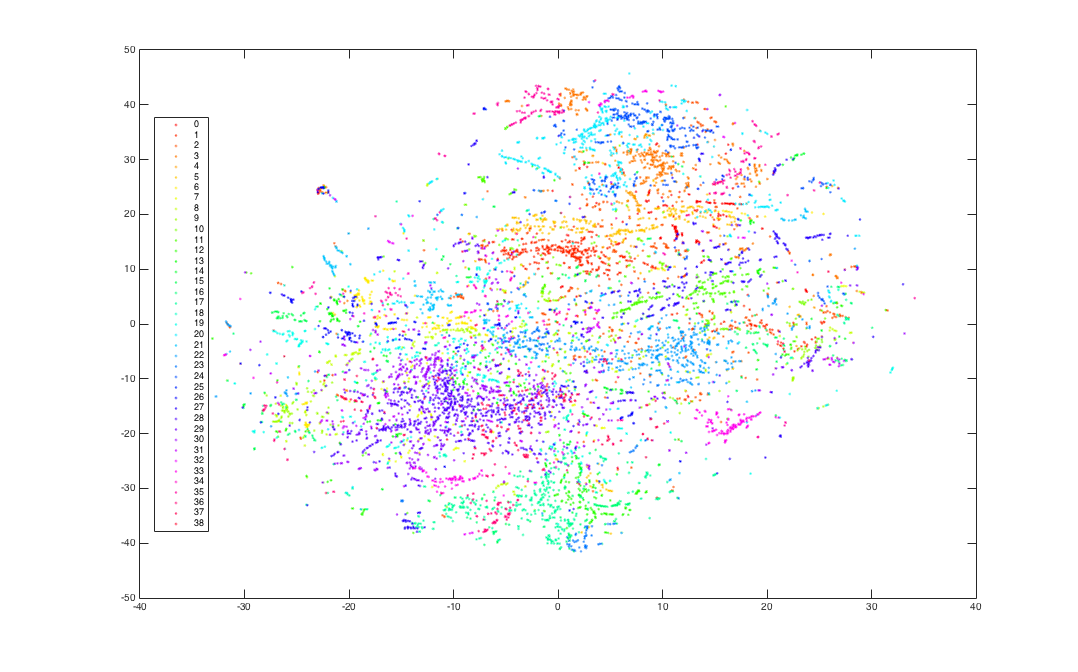}
\includegraphics[height=2in, width=2.3in]{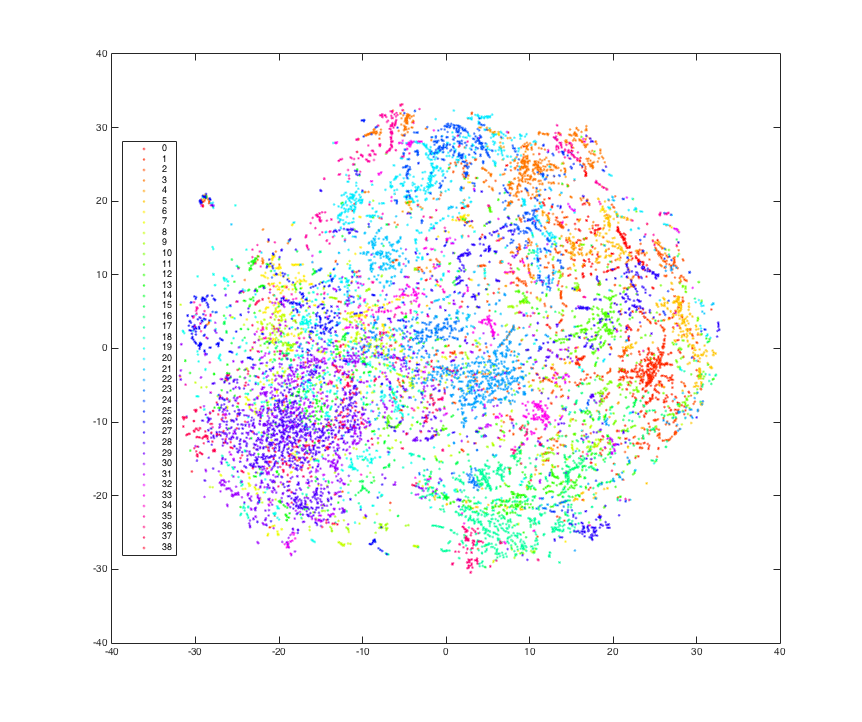}
\includegraphics[height=2in, width=2.3in]{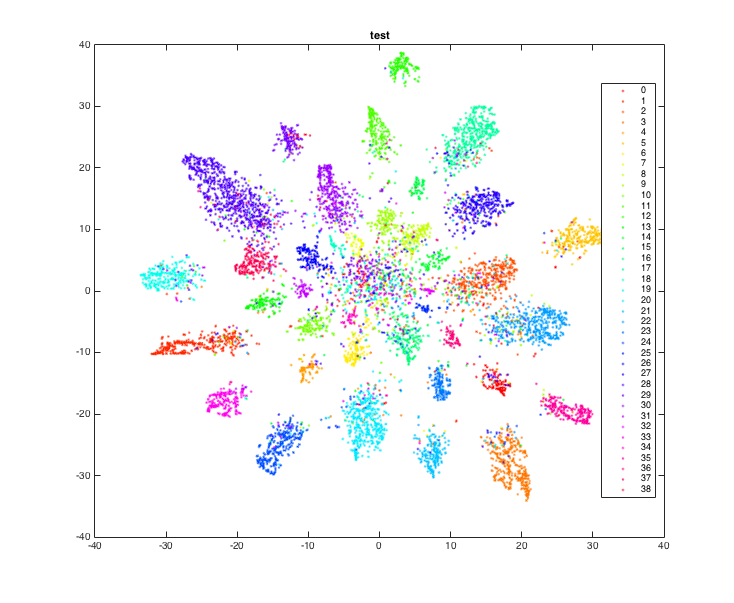}
\caption{Phoneme clusters as computed via CCA, KCCA, and SplitAE}
\label{fig:tsne}
\end{figure*}

For CCA, KCCA, and SplitAE we computed clusters using t-SNE in order to see how well the three methods preserve phoneme clusters. We have presented the three cluster charts in \ref{fig:tsne} based on the test data that was transformed by the aforementioned methods. Using the same test data, we performed k-NN classification and give the results in Table \ref{tbl:classification}. For KCCA, we did not use approximation methods in order to compute the kernel matrices and so only provide this analysis for the sake of completeness. It is computationally intractable to compute kernel matrices on large datasets without approximation methods (such as Fourier features or Nystrom approximation). This is the reason why KCCA performs worse at classification as compared to the baseline, CCA, and SplitAE. 

As we can see, however, correlation methods perform better than the baseline. In addition, SplitAE shows the power of deep neural networks in unsupervised learning and so it is clear that deep neural networks trained with correlation (DCCA and DCC-LSTM) would outperform CCA, KCCA, and SplitAE.

\subsection{Reconstruction}

\begin{table}[]
\centering
\begin{tabular}{lll}
 & Baseline & DCC-LSTM \\
Correlation Captured                                                 & 15.1242            & 24.2                                     \\
Correlation of Top 20 Components                                     & 7.6268             & 11.2986                                  \\
Sum of Distances to Nearest Neighbors                                & 15,778             & 3,296                                    \\
Reconstruction Error (L2 norm)                                       & 43,949             & 26,285                                   \\
Per-Sample Reconstruction Error                           & 4.395              & 2.6286                                   \\
** Per-Vector-Per-Component Error                     & 0.21975            & 0.13143                                 
\end{tabular}
\caption{Nearest Neighbor Reconstruction Task: reconstruct articulatory test data from acoustic test data in 20-dim shared CCA space (10k random test samples)}
\label{tbl:reconstruction}
\end{table}

The downstream speakers were further split into a 10,000 sample test set, while the remaining were train. Linear CCA was performed on the feature space representations of the train samples, projecting both of their views into a shared CCA space. Because articulatory information is rarely available at test time, we reconstructed the 10,000 articulatory vectors from their nearest neighbors in the CCA space and recorded the average $\ell_2$ reconstruction error. 

Table \ref{tbl:reconstruction} shows the comparison between baseline reconstruction and DCC-LSTM reconstruction. As we can see, the data transformed by DCC-LSTM significantly improves on the reconstruction error. Additionally, we can see from this table that the top twenty correlated components capture about half of the total correlation.

\section{CONCLUSION}

It's not immediately clear that DCCA and DCC-LSTM are comparable. DCC-LSTM requires contiguous frames to learn, and so randomization must be done with sequences whereas DCCA randomizes the frames it learns from in batches. XRMB data has lots of spaces/pauses which end up having high correlation as nothing is happening in the articulatory feature space when there is a space. We've seen that DCCA performs well when trained on data that has spaces/pauses removed, whereas DCC-LSTM cannot be trained on such a dataset. 

With fifty output dimensions and one hidden layer, DCCA is able to capture $35 \pm 5$ correlation per speaker (view 1 and view 2 hidden widths of 1800 and 1200, respectively). DCC-LSTM is able to capture $27 \pm 3$ correlation per speaker (view 1 and view 2 hidden widths of 400 each). It's clear that DCC-LSTM is able to capture correlation, and be trained in this unsupervised manner. In addition, we are able to capture quite a bit of correlation with significantly less parameters/network width and depth. This is an indication that with enough parameter tuning and training, DCC-LSTM has the potential to outperform DCCA on time-series multi-view data.

There is much work to be done still, in figuring out the efficacy of DCC-LSTM. We hope to have given enough proof in this paper to warrant exploring this method further. Current attempts to train LSTMs in an unsupervised fashion use reconstruction as the objective, nobody has attempted to use correlation as of yet. This is an important result as we see that it is possible to use correlation. In addition, there is an extension of DCCA called DCCAE (Deep Canonically Correlated AutoEncoders) in which the correlation objective is added to the reconstruction objective. It is seen to do much better than DCCA, so the natural extension for DCC-LSTM would be to add the sequence reconstruction or sequence prediction objective into the correlation objective.

\addtolength{\textheight}{-12cm}   









\bibliographystyle{IEEEtran}
\bibliography{scholar}

\end{document}